\documentclass[runningheads]{article}
\usepackage{graphicx}
\usepackage{subfigure}
\usepackage{amsmath,amssymb} 
\usepackage{bm}
\usepackage[justification=centering]{caption}
\usepackage{color}
\usepackage{textcomp}
\usepackage{multirow}
\usepackage{float}
\usepackage[top = 12mm]{geometry}
\begin{document}
	
	\title{3D Keypoint Detection Based on Deep Neural Network with Sparse Autoencoder} 
	
	\author{Xinyu Lin, Ce Zhu, Qian Zhang, Yipeng Liu}
	\date{}
	\renewcommand{\thefootnote}{\fnsymbol{footnote}}
	\footnotetext[1]{All the authors are with School of Electronic Engineering / Center for Robotics, University of Electronic Science and Technology of China (UESTC), Chengdu, China. (email: xinyulin@std.uestc.edu.cn; eczhu@uestc.edu.cn; zhangqian@std.uestc.edu.cn; yipengliu@uestc.edu.cn)}
	
	\maketitle

\begin{abstract}
Researchers have proposed various methods to extract 3D keypoints from the surface of 3D mesh models over the last decades, but most of them are based on geometric methods, which lack enough flexibility to meet the requirements for various applications. In this paper, we propose a new method on the basis of deep learning by formulating the 3D keypoint detection as a regression problem using deep neural network (DNN) with sparse autoencoder (SAE) as our regression model. Both local information and global information of a 3D mesh model in multi-scale space are fully utilized to detect whether a vertex is a keypoint or not. SAE can effectively extract the internal structure of these two kinds of information and formulate high-level features for them, which is beneficial to the regression model. Three SAEs are used to formulate the hidden layers of the DNN and then a logistic regression layer is trained to process the high-level features produced in the third SAE. Numerical experiments show that the proposed DNN based 3D keypoint detection algorithm outperforms current five state-of-the-art methods for various 3D mesh models.
\end{abstract}

\section{Introduction}
Detection of 3D keypoints has been a very popular approach within 3D computer vision for various applications, such as object registration \cite{Registration2005robust}, 3D shape retrieval \cite{Retrieval2004shape}, object matching \cite{hu2009salient}, mesh segmentation \cite{segmentation2005mesh} and simplification \cite{Meshsaliency2005mesh}. Researchers have proposed various methods to extract 3D keypoints from the surface of 3D mesh models over the last decades. Most of 3D keypoint detection algorithms are based on geometric methods \cite{Meshsaliency2005mesh,3Dsift2011salient,3Dharris2011harris,sun2009concise,novatnack2007scale,don20143d,castellani2008sparse,wang2013sdtp,akagunduz2009scale,meshdog2009,LSP20073d,ISS2009,KPS2010repeatability,LBSS2008,scakemeshdog}. Godila and Wagan \cite{3Dsift2011salient} proposed a method for detecting the 3D salient local features on the  basis of voxel grid inspired by the Scale Invariant Feature Transform (SIFT) algorithm \cite{sift2004distinctive}. Sipiran and Bustos \cite{3Dharris2011harris} proposed an effective and efficient extension of the Harris operator \cite{harris1988combined} for 3D objects. Lee et al. \cite{Meshsaliency2005mesh} defined mesh saliency in a scale-dependent manner utilizing a center-surround operator on Gaussian-weighted mean curvatures and used it as a measure of regional importance for 3D mesh models. Holte utilized Difference-of-Normals operator to address the problem of detecting 3D keypoints \cite{don20143d}. Castellani et al. \cite{castellani2008sparse} proposed a salient point detection algorithm where sparse 3D keypoints are selected robustly by exploiting visual saliency principles on 3D mesh models. Besides of the methods mentioned above, there are other methods based on Laplacian spectrum \cite{hu2009salient,song20133d,song2014mesh}, which extract salient geometric feature points in Laplace-Beltrami spectral domain instead of spatial domain.

As is described in \cite{randomforest20143d}, using geometric methods to detect 3D keypoints lacks enough flexibility to meet the requirements for various applications because of the following three reasons: 1) Different tasks have different requirements for 3D keypoint detection algorithm: high recall is necessary in some tasks while others may require high precision \cite{mikolajczyk2005performance}. 2) Geometric methods usually assume that the vertices with sharp changes in the 3D models are 3D keypoints, but in fact, these vertices may be noise or local variation. 3) Using geometric methods encounters various difficulties when semantic ambiguity is considered. All of these reasons drive researchers to find a new framework to detect 3D keypoints. 

In recent years, some researchers proposed 3D keypoint detection algorithms based on machine learning \cite{randomforest20143d,machine3dkeypoint2013machine,salti2015learning}, which could solve these problems mentioned above to some extent. But most of them only utilized local information to detect 3D keypoints, lacking corresponding global information such as Laplacian spectrum. Teran and Mordohai \cite{randomforest20143d} proposed a 3D keypoint detection algorithm using a random forest \cite{ran2001random} as the classifier, where several geometric detectors are used to produce attributes. Creusot et al. \cite{machine3dkeypoint2013machine} utilized a linear method, namely Linear Discriminant Analysis and a non-linear method, namely AdaBoost \cite{adaboost1997decision} to detect 3D keypoints from 3D face scans. Salti and Tombari \cite{salti2015learning} cast 3D keypoint detection as a binary  classification between points whose support can be correctly matched by a pre-defined 3D descriptor (SHOT descriptor \cite{shot2010unique}) or not, and the same with \cite{randomforest20143d}, random forest \cite{ran2001random} is used as the classifier.

In this paper, we propose a new 3D keypoint detection algorithm on the basis of deep learning. Here we formulate the 3D keypoint detection as a regression problem using deep neural network (DNN) with sparse autoencoder (SAE) \cite{sae2011sparse} as our regression model. Both local information and global information of a 3D mesh model in multi-scale space are fully utilized to detect whether a vertex is a keypoint or not. SAE can effectively extract the internal structure of these two kinds of information and formulate high-level features for them, which is beneficial to the regression model. Three SAEs are used to formulate the hidden layers of the DNN and then a logistic regression \cite{logistic1958regression} layer is trained to process the high-level features produced in the third SAE. These four layers are stacked together to formulate a DNN as the regression model of the proposed 3D keypoint detection algorithm. Numerical experiments demonstrate that the DNN approach proposed by us outperforms the existing five state-of-the-art methods for various 3D mesh models. 
 
The main contributions of this paper are summarized as follows: (a) It's the first time that DNN with SAE has been used to detect 3D keypoints. (b) Both local information and global information of a 3D mesh model in multi-scale space are fully utilized to detect whether a vertex is a keypoint or not. (c) Numerical experiments on the datasets \cite{dutagaci2012evaluation} are presented to verify the performance of the proposed DNN based 3D keypoint detection algorithm.

The rest of this paper is organized as follows. Introduction of DNN with SAE is described in Section 2. Section 3 presents the attributes and the training process. The proposed DNN based 3D keypoints detection algorithm will be displayed in Section 4 and performance study and result analysis are shown in Section 5. Finally, conclusions are presented in Section 6.

\section{Deep Neural Network with Sparse Autoencoder}
\subsection{Sparse Autoencoder}
Autoencoder \cite{autoencoder2006reducing} is an unsupervised learning algorithm, in which target values are set to be equal to the input values. It tries to learn the function \emph{$\bm{h_{W,b}}\approx \bm{x}$}, where \emph{$\bm{x}$} represents the unlabeled dataset \emph{$\{x_1,x_2,...,x_m\},x_i \in \mathbb{R}^m$}. The left part of Fig. 1 displays the framework of an autoencoder, in which the hidden layer (layer 2 in Fig. 1) contains the internal structure of input data.  Similar to Principal Component Analysis \cite{pca2002principal} (PCA), autoencoder can learn a low-dimensional representation of the input data by limiting the number of hidden units.
\begin{figure}[tbp]
	\centering
	\begin{minipage}[b]{0.95\linewidth}
		\includegraphics[width=1\textwidth]{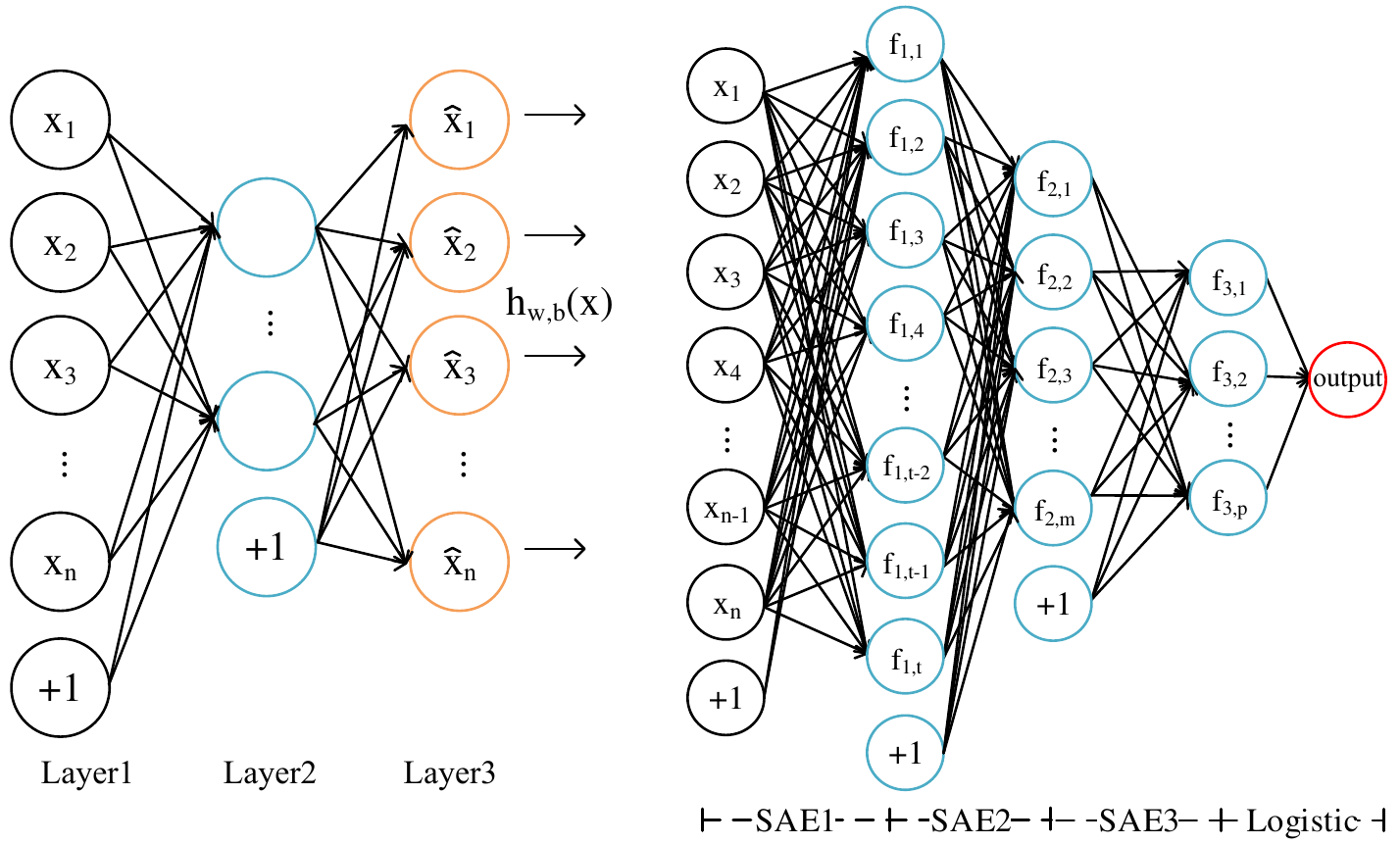}
		\caption{Left: the framework of an autoencoder. Right: DNN with three SAEs as the hidden layers and a logistic regression layer}
	\end{minipage}
\end{figure}

Sparse autoencoder \cite{sae2011sparse,SAE2009_3790} is a variant of autoencoder, which can be formed by imposing a sparsity constraint on the hidden units. It can also learn interesting structure of the input data, even if the dimension of hidden layer is larger than the dimension of input layer. The overall cost function of sparse autoencoder is
\begin{equation}
	J_{sparse}(\bm{W,b}) = \frac{1}{2}||\bm{h_{W,b}}-\bm{x}||^{2} + \beta \sum_{j=1}^{S}\text{KL}(\rho||\widehat{\rho_j})\\
\end{equation}
where
\begin{gather}
\text{KL}(\rho||\widehat{\rho_j}) = \rho \log \frac{\rho}{\widehat{\rho_j}} + (1-\rho)\log \frac{1-\rho}{1-\widehat{\rho_j}}\\
\widehat\rho_j = \frac{1}{m} \sum_{i=1}^{m}[a_j^2(x_i)]
\end{gather}
where \emph{$a_j^2(x)$} denotes the activation of hidden unit \emph{$j$} when the network is given a specific input \emph{$x$} and \emph{$\widehat\rho_j$} represents its average activation. The second term of (1) is a penalty term which penalizes \emph{$\widehat\rho_j$} deviating significantly from \emph{$\rho$} and (approximately) enforce the constraint \emph{$\widehat{\rho_j} = \rho$}, where \emph{$\rho$} is a sparsity parameter, typically a small value close to zero. KL\emph{$(\rho||\widehat{\rho_j})$} is the Kullback-Leibler divergence \cite{kl1951information} between a Bernoulli random variable with mean \emph{$\rho$} and a Bernoulli random variable with mean \emph{$\widehat\rho_j$}, and it is a standard function for measuring the level of difference between two different distributions. \emph{$\beta$} controls the weight of the sparsity penalty term. \emph{$S$} is the number of neurons in the hidden layer.

\subsection{Deep Neural Network with Sparse Autoencoder}
Neural networks with multiple hidden layers can be useful for solving classification and regression problems with complex data. Each layer can learn features at different level of abstraction. However, there exist some problems when training neural networks with multiple hidden layers in practice, e.g. the local minimum problem of weights.  

In this paper, we select SAE to formulate the hidden layers of DNN for the following three reasons. The first reason is its excellent performance according to \cite{sae2011sparse,SAE2009_3790}. SAE can learn interesting structure of the input data effectively if there exists any correlation in the input data. Besides, SAE has great ability for feature processing according to \cite{icasspsae}. Using deep sparse autoencoder (DSAE, stacked by a few successive SAEs) can learn high-level features of the input data effectively. Each SAE in DSAE can learn features at different levels (from low level to high level). In addition, the process of training SAEs can be viewed as a stage of pre-training of DNN, which makes the initial weights of DNN close to the position of global optimization. So, it can effectively avoid the local minimum problem of weights mentioned in the last paragraph.

To formulate the DNN regression model, we firstly train three SAEs and select the encoder part to formulate the hidden layers of the DNN. As is shown in right part of Fig. 1, the input of the first SAE is the original input data and its hidden layer is regarded as input of the second SAE. Likewise, the hidden layer of second SAE is regarded as the input of the third SAE. This process can be viewed as a stage of high-level features formulation according to \cite{icasspsae}. Then we train a logistic regression layer to process the high-level features produced in the third SAE. Finally, we stack the four layers together to formulate a DNN as our regression model.

\section{Attributes and Training Process}
For 3D mesh models, there is no additional information other than the position of vertices and the connectivity information among these vertices. If we have enough 3D mesh models with ground truth, we can directly utilize these information to train the DNN regression model. However, the training data in 3D mesh models with ground truth, are limited. So, to improve the performance of DNN regression model, what we first need to do is preprocessing the original data to formulate the attributes as the inputs to our DNN regression model.

\subsection{Attributes}
In this section, we utilize both local and global information of 3D mesh models in multi-scale space to formulate the attributes as the inputs to our DNN regression model. For local information, we utilize three types of geometric properities of surface of a 3D mesh model: (1) the Euclidean distance between neighborhood rings to the tangent plane; (2) the angles of normal vectors between the vertex and its neighborhood rings; (3) various curvatures. For global information, we consider the properties of log-Laplacian spectrum of a 3D mesh model \cite{song2014mesh}. For any vertex \emph{$v$} in a 3D mesh model \emph{$\bm{M}(x,y,z)$}, let \emph{$\bm{f}$} be its attribute. It can be formulated by 
\begin{gather}
	\bm{f} = [\bm{f}_0,\bm{f}_1,\bm{f}_2,...,\bm{f}_\varOmega]^T\\
	\bm{f}_i = [\bm{f}_{d},\bm{f}_{\theta},\bm{f}_{c},\bm{f}_{ls}], i = 0,1,2,...,\varOmega
\end{gather}
where \emph{$\bm{f}_i,i=0,1,2,...,\varOmega$} represents the information in scale \emph{$i$} of a 3D mesh model which can be denoted as \emph{$\bm{M}_\delta(x,y,z)$}. \emph{$\bm{M}_\delta(x,y,z)$} can be calculated by
\begin{gather}
	\bm{M}_\delta(x,y,z)=\bm{M}(x,y,z)*\bm{G}(x,y,z,\delta)\\
	\bm{G}(x,y,z,\delta)=\frac{1}{(\sqrt{2\pi}\delta)^{3}}e^{-\frac{(x^{2}+y^{2}+z^{2})}{2\delta^2}}
\end{gather}
where \emph{$\delta \in \{0,\varepsilon,2\varepsilon,...,\varOmega\varepsilon\}$} is the standard deviation of 3D Gaussian filter and \emph{$\varepsilon$} amounts to 0.3\% of the length of the main diagonal located in the bounding box of the model. \emph{$\bm{M}_0(x,y,z)$} indicates that it is the original mesh model \emph{$\bm{M}(x,y,z)$}. \emph{$*$} is the convolution operator.

\emph{$\bm{f}_i$} is made up of  four parts: three types of local information (\emph{$\bm{f}_d,\bm{f}_\theta,\bm{f}_c$}) and one type of global information (\emph{$\bm{f}_{ls}$}).

\subsubsection{Local Information}
According to the second fundamental form of a surface \cite{harris1989second}, the most direct measurement which indicates the curvature degree of a vertex \emph{$v$} on the surface of a 3D mesh model is the Euclidean distance between neighborhood vertices to the tangent plane of the vertex \emph{$v$}. The angles of normal vectors between the vertex and its neighborhood rings is another important geometric property for the surface of a 3D mesh model \cite{don20143d}.

Here, we calculate the first two types of local information \emph{$\bm{f}_d$} and \emph{$\bm{f}_\theta$}. In this paper, a 3D mesh model (triangular mesh models of arbitrary topology) is represented as a set of vertices \emph{$\bm{V}$} and faces \emph{$\bm{F}$} with adjacency information between these entities. For different vertices in a 3D mesh model, they may have different numbers of neighborhood vertices. We utilize an adaptive technique in \cite{3Dharris2011harris} to find the neighborhood vertices of a vertex. Let \emph{$v$} be the analyzed vertex and \emph{$\bm{V}_k(v),k=1,2,3,4,5$} be the \emph{$k$}-ring neighborhood vertices around \emph{$v$}. As is shown in Fig. 2, bule dots, megenta dots, green dots, cyan dots and yellow dots represent the first, second, third, fourth and fifth ring around vertex \emph{$v$} respectively.  For any vertex \emph{$v$} in a 3D mesh model, we utilize \emph{$\bm{n}$} to represent its normal vector. Let \emph{$v_{kj}$} be the \emph{j}-th point in \emph{$\bm{V}_k(v)$} and \emph{$\bm{n}_{kj}$} be its normal vector. Let \emph{$d_{kj}$} be the Euclidean distance of \emph{$v_{kj}$} to the tangent plane and \emph{$\theta$} be the angle of normal vector between \emph{$v$} and \emph{$v_{kj}$}, both of which can be calculated by 
\begin{figure}[tbp]
	\centering
	\begin{minipage}[b]{1\linewidth}
		\includegraphics[width=1\textwidth]{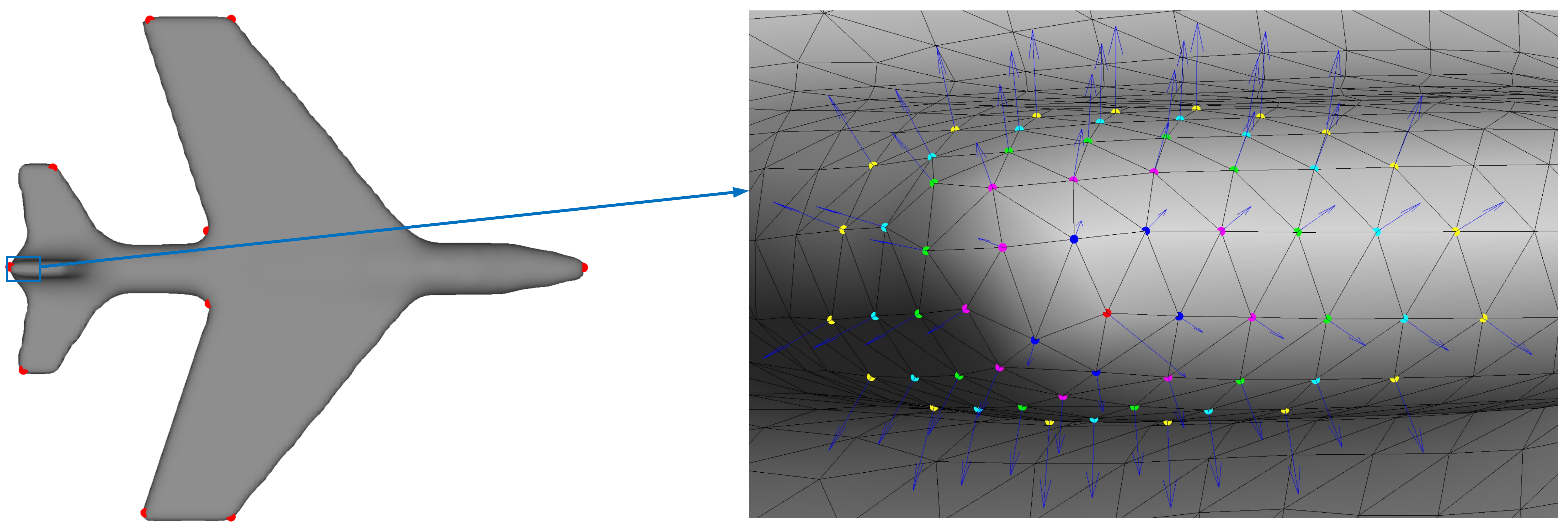}
		\caption{Left: airplane mesh model with ground truth; Right: part of airplane mesh model. (Vertex \emph{$v$} (red dot) and its neighborhood rings: \emph{$\bm{V}_1(v)$}, blue dots; \emph{$\bm{V}_2(v)$}, magenta dots; \emph{$\bm{V}_3(v)$}, green dots; \emph{$\bm{V}_4(v)$}, cyan dots; \emph{$\bm{V}_5(v)$}, yellow dots.)}
	\end{minipage}
\end{figure}
\begin{gather}
	d_{kj} = \frac{|\bm{n}^T[x_{kj},y_{kj},z_{kj}]^T-\bm{n}^T[x_v,y_v,z_v]^T|}{|| \bm{n} ||_2}\\
	\theta_{kj}= \min(\arccos(\frac{\bm{n}^T\bm{n}_{kj}}{||\bm{n}||_2||\bm{n}_{kj}||_2})
\end{gather} 
where \emph{$(x_v,y_v,z_v)$} is the coordinate of \emph{$v$} and \emph{$(x_{kj},y_{kj},z_{kj})$} is the coordinate of \emph{$v_{kj}$}. Let \emph{$\bm{d}_k$} be \emph{$[d_{k1},d_{k2},...,d_{kN_k}]^T$} and \emph{$\bm{\theta}_k$} be \emph{$[\theta_{k1},\theta_{k2},...,\theta_{kN_k}]^T$}, where \emph{$N_k$} is the number of vertices in \emph{$\bm{V}_k(v)$}. Because there are different numbers of neighborhood vertices for different vertices in a 3D mesh model, we utilize six types of statistical properties of \emph{$\bm{d}_k$} and \emph{$\bm{\theta}_k$} to formulate \emph{$\bm{f}_d$} and \emph{$\bm{f}_\theta$} respectively in order to form the fixed dimension of attributes. \emph{$\bm{f}_d$} and \emph{$\bm{f}_\theta$} can be calculated by
\begin{gather}
	\begin{split}
		\bm{f}_d = [\max(\bm{d}_k),\min(\bm{d}_k),\max(\bm{d}_k)-\min(\bm{d}_k),\\
		\text{mean}(\bm{d}_k),\text{var}(\bm{d}_k),\text{harmmean}(\bm{d}_k)] \ \ \ \ \ \ \ 
	\end{split}\\
	\begin{split}
		\bm{f}_\theta = [\max(\bm{\theta}_k),\min(\bm{\theta}_k),\max(\bm{\theta}_k)-\min(\bm{\theta}_k),\\
		\text{mean}(\bm{\theta}_k),\text{var}(\bm{\theta}_k),\text{harmmean}(\bm{\theta}_k)]  \ \ \ \ \ \ \ 
	\end{split}
\end{gather}
where \text{mean}\emph{$(\bm{\cdot})$}, \text{var}\emph{$(\bm{\cdot})$} and \text{harmmean}\emph{$(\bm{\cdot})$} are the arithmetic average, variance and harmonic average  respectively. 

As is shown in Table 1, for any vertex \emph{$v$} located in flat area of a 3D mesh model, all six types of statistical properties of \emph{$\bm{d}_k$} and \emph{$\bm{\theta}_k$} should be small. For any vertex \emph{$v$} located in edges of a 3D mesh model, \text{max}\emph{$(\bm{\cdot})$}, \emph{$\max(\bm{\cdot})-\min(\bm{\cdot})$} and \text{var}\emph{$(\bm{\cdot})$} should be large. However, \text{min}\emph{$(\bm{\cdot})$} and \text{var}\emph{$(\bm{\cdot})$} should be small. For any vertex \emph{$v$} belonging to 3D keypoints of a 3D mesh model, \emph{$\max(\bm{\cdot})$}, \emph{$\min(\bm{\cdot})$}, \text{mean}\emph{$(\bm{\cdot})$} and \text{harmmean}\emph{$(\bm{\cdot})$} should be large. However, \emph{$\max(\bm{\cdot})-\min(\bm{\cdot})$} and \text{var}\emph{$(\bm{\cdot})$} should be small.

\begin{table}[tbp]  
	\centering
	\renewcommand\arraystretch{1.2}
	\caption{The effect of six types of statistical properties presented on different regions of a 3D mesh model}
	\begin{tabular}{ccccccc}  
		\hline
		\cline{1-7}
		\ &\emph{$\max(\cdot)$} \ \ &\emph{$\min(\cdot)$} \ \  &\emph{$\max(\cdot)-\min(\cdot)$}  \ \ &\text{mean}\emph{$(\bm{\cdot})$} \ \ &\text{var}\emph{$(\bm{\cdot})$} \ \ &\text{harmmean}\emph{$(\bm{\cdot})$}\\ \hline  
		Flat areas &small &small &small &small &small &small\\        
		Edges &large &small &large &relatively large &large &small\\        
		3D keypoints &large &large &small &large &small &large\\        
		\hline
		\cline{1-7}
	\end{tabular}
\end{table}

Besides of the two geometric properties mentioned above, curvatures \cite{Curvatures} are frequently used to detect saliency of 3D mesh models according to \cite{Meshsaliency2005mesh,akagunduz2009scale,LSP20073d,randomforest20143d,machine3dkeypoint2013machine}. In this paper, four types of curvatures are used to formulate the \emph{$\bm{f}_c$} and it can be formulated as
\begin{equation}
	\bm{f}_c = [c_1,c_2,\frac{c_1+c_2}{2},c_1c_2]
\end{equation}
where \emph{$c_1$} and \emph{$c_2$} are principal curvatures. \emph{$(c_1+c_2)/2$} is mean curvature, and \emph{$c_1c_2$} is gaussian curvature.

\subsubsection{Global Information}
As a powerful tool, Laplacian spectrum is widely used to analyze global information of 3D mesh models according to \cite{hu2009salient,song20133d,song2014mesh,Pauly,Levy:2010:SMP:1837101.1837109,Zhang:2012:VMD:2167076.2167079}. The Laplacian matrix \emph{$\bm{L}$} of a 3D mesh model is a symmetric matrix and it can be decomposed as:
\begin{equation}
\bm{L} = \bm{B \wedge B}^T
\end{equation}
where \emph{$\bm{\wedge} = \text{\emph{Diag}}\{\lambda_f,1\leqslant f \leqslant \Psi\}$} is a diagonal matrix arranged in ascending order, in which \emph{$\lambda_f$} is the eigenvalue of \emph{$\bm{L}$}. The corresponding eigenvectors are utilized to form the orthogonal matrix \emph{$\bm{B}$}. \emph{$\Psi$} is the number of vertices of a 3D mesh model. \emph{$\bm{H}(f) = \{\lambda_f,1\leqslant f \leqslant \Psi\}$} is the Laplacian spectrum of a 3D mesh model.

In this paper, we get global information via log-Laplacian spectrum used in \cite{song2014mesh}, which is defined as
\begin{equation}
	\bm{L}(f) = \log(|\bm{H}(f)|)
\end{equation}
Spectral irregularity \emph{$\bm{R}$} is utilized to calculate the mesh saliency and it can be formulated as 
\begin{equation}
	\bm{R}(f) = |\bm{L}(f)-\bm{J}_\Gamma(f)*\bm{L}(f)|
\end{equation}
where \emph{$\bm{J}_\Gamma(f) = \frac{1}{\Gamma}[1,1,...,1]$} is \emph{$1\times\Gamma$} vector. To bring this representation back to the spatial domain, the composition should be formulated as:
\begin{equation}
	\bm{S} = \bm{B}\bm{R}_1\bm{B}^T\odot \bm{W}
\end{equation}
where \emph{$\bm{R}_1 = \emph{Diag}\{\emph{exp}(\bm{R}(f)):1\leqslant f\leqslant \Psi\}$} is a diagonal matrix. \emph{$\bm{W}$} is the distance-weighted adjacency matrix and \emph{$\odot$} is Hadamard product. 

Let \emph{$\bm{s}_k$} be \emph{$[s_{k1},s_{k2},...,s_{kN_k}]^T$} and \emph{$s$} be the element of \emph{$\bm{S}$}. The same with last section, we also utilize the six types of statistical properties of \emph{$\bm{s}_k$} to formulate \emph{$\bm{f}_{ls}$}. It can be calculated by 
\begin{equation} 
	\begin{split}
		\bm{f}_{ls} = [\max(\bm{s}_k),\min(\bm{s}_k),\max(\bm{s}_k)-\min(\bm{s}_k),\\
		\text{mean}(\bm{s}_k),\text{var}(\bm{s}_k),\text{harmmean}(\bm{s}_k)] \ \ \ \ \ \ \ 
	\end{split}
\end{equation}

\subsection{Training Process}
We use the same datasets as in \cite{song20133d,randomforest20143d,dutagaci2012evaluation} where a web-based application is developed and utilized to collect ground truth of 3D keypoints on 43 mesh models. These mesh models are organized in two datasets. The first one (Dataset A) is constituted by 24 triangular mesh models and annotated by 23 human subjects. Another one (Dataset B) is constituted by 43 triangular mesh models and annotated by 16 human subjects. Similar to \cite{randomforest20143d}, for all the experiments, we select two-thirds of Dataset A and two-thirds of Dataset B to train our DNN regression model. The remained mesh models are used as test datasets and they are never used in training the DNN regression model. As is shown in Table 2, for Dataset A, the representative of clusters \emph{$\sigma \in \{0.01,0.02,...,0.1\}$} and \emph{$n\in \{11,12,...,22\}$} are placed in the positive class. For Dataset B, the representative of clusters \emph{$\sigma \in \{0.01,0.02,...,0.1\}$} and \emph{$n\in \{8,9,...,15\}$} are placed in the positive class. All other vertices are placed in the negative class.

\begin{table}[tbp]  
	\centering
	\renewcommand\arraystretch{1.2}
	\caption{Training datasets}
	\begin{tabular}{cccccc}  
		\hline
		\cline{1-6}
		Datasets \ &Mesh models \ &\emph{$n$} \ &\emph{$\sigma$} \ \  &Positive samples \ &Negative samples \\ \hline  
		A &16 &\emph{$\{11,...,22\}$} \ \  &all &17115 &148565\\        
		B &28 &\emph{$\{8,...,15\}$} \ \ &all &18427 &222034\\        
		\hline
		\cline{1-6}
	\end{tabular}
\end{table}
\begin{table}[tbp]  
	\centering
	\renewcommand\arraystretch{1.2}
	\caption{Parameters for DNN regression model}
	\begin{tabular}{ccccc}  
		\hline
		\cline{1-5}
		\ \ \ &Dimension of input layer \ \ \ &Dimension of hidden layer \ \ \ \ &\emph{$\rho$} \ \ \ \  &\emph{$\beta$} \\ \hline  
		SAE1 &665 &800 &0.15 \ \ \ &4 \\               
		SAE2 &800 &200 &0.15 \ \ \ &4 \\               
		SAE3 &200 &50 &0.1 \ \ \ &4 \\               
		Logisitc &50 &1 & \ \ \ & \\    
		\hline
		\cline{1-5}
	\end{tabular}
\end{table}

To train the DNN regression model, we need to train three SAEs firstly. All the parameters related to the three SAEs are displayed in Table 3. Then a logistic regression layer is trained. Finally, we stack the four layers together to formulate a DNN as our regression model and fine tuning is done by performing backpropagation on the DNN to improve the performance of DNN regression model.

\section{Proposed DNN Based 3D Keypoint Detection Algorithm}
The outline of the proposed DNN based 3D keypoint detection algorithm can be divided into the following four steps:
\begin{itemize}
	\item Utilize 3D Gaussian filter to construct scale space for a 3D mesh model \emph{$\bm{M}(x,y,z)$} and get a series of evolved 3D mesh models \emph{$\bm{M}_\delta(x,y,z)$}.
	\item Use multi-scale information to calculate the attributes for every vertex of a 3D mesh model in the same way as described in Section 3.
	\item For every vertex of a 3D mesh model, put its attributes to the well-trained DNN regression model and get a regression value, and then get the saliency map of this 3D mesh model.
	\item Select the local maxima of saliency map of a 3D mesh model as the 3D keypoints. Compare the value of DNN regression \emph{$\rho$}  for every vertex with those points in its neighborhood rings \emph{$V_k(v),k=1,2,3,4,5$} and select the maximal one as the 3D keypoints. 
\end{itemize}

\section{Numerical Experiments}

In this section, we evaluate the performance of DNN based 3D keypoint detection algorithm and compare it with five state-of-the-art methods - namely 3D Harris \cite{3Dharris2011harris}, HKS \cite{sun2009concise}, Salient Points \cite{castellani2008sparse}, Mesh Saliency \cite{Meshsaliency2005mesh} and Scale Dependent Corners \cite{novatnack2007scale}. All of the five methods are referenced algorithms for performance evaluation in \cite{song20133d,randomforest20143d,dutagaci2012evaluation}.

\subsection{Datasets and Evaluation Metrics}

As is described in section 3.2, we utilize the remained one-third of Dataset A and one-third of Dataset B as the test datasets. Some evaluation methods usually measured the repeatability rate according to varying factors, such as model deformation, scale change, different modalities, noise, and topological change \cite{tombari2013performance}. Different from them, Dutagaci et al. utilized three evaluation metrics - namely False Positive Error (FPE), False Negative Error (FNE) and Weighted Miss Error (WME) to evaluate the performance of 3D keypoints detection algorithms \cite{dutagaci2012evaluation}.

Besides of the evaluation metrics mentioned above, Teran \cite{randomforest20143d} also adopted the Intersection Over Union (IOU) [29] as their main metric to evaluate the performance of the 3D keypoint detection algorithms. It can be calculated by
\begin{equation}
	IOU(r) = \frac{TP}{FN+FP+TP}
\end{equation}
where \emph{$FP = N_A-N_C$} is the number of false positives and \emph{$FN = N_G-N_C$} represents the number of false negatives. \emph{$TP = N_C$} is the number of true positives. \emph{$N_G$} is the number of ground truth points, \emph{$N_C$} is the number of correctly detected points and \emph{$N_A$} denotes the number of detected 3D keypoints by the algorithm. \emph{$r$} is localization error tolerance \cite{dutagaci2012evaluation}.

Similar to Dutagaci et al. \cite{dutagaci2012evaluation}, Song et al. \cite{song20133d} and Teran et al. \cite{randomforest20143d}, we utilize the same two datasets to evaluate the performance of six algorithms in terms of four evaluation metrics - namely IOU, FNE, FPE and WME. The IOU is chosen as the main metric because FNE and FPE can be misleading in isolation according to the discussion in \cite{randomforest20143d}. Besides, a 3D video sequence produced in \cite{3Dsequence} is used to verify the stablity of the proposed DNN based 3D keypoints detection algorithm.

\subsection{Experimental Results}
In this section, we present the performance results of the proposed DNN based 3D keypoint detection algorithm and other five state-of-the-art 3D keypoint detection algorithms. In all the experiments, \emph{$\Omega$} is 6 and \emph{$\Gamma$} is 9. It is important to note that the 3D keypoints detected by the six 3D keypoint detection algorithms are constant when \emph{$n/\sigma$} varies, but the ground truth is variable when \emph{$n/\sigma$} varies according to \cite{dutagaci2012evaluation}. 

Fig. 3 displays two types of visualizations of the results. Fig. 3(a) displays the 3D keypoints of chair model detected by our proposed DNN based 3D keypoint detection algorithm (second row) and corresponding saliency maps (first row) from four different viewpoints. Fig. 3(b) displays 3D keypoints and corresponding saliency maps of five frames (the 1st, 25th, 50th, 75th and 100th frame respectively) which come from a 3D video sequence \cite{3Dsequence} from the same viewpoint. Visualizations of comparative results can be found from Fig. 4, where 3D keypoint of chair model in Dataset A and armadillo model in Dataset B are detected by six methods. More visualizations of the results can be found in supplementary material. From the visualizations of the results, we can see that the 3D keypoints detected by our approach are more in accord with human visual characteristics. Besides, the 3D keypoints detected by our approach are stable according to Fig. 3(b), where the distribution and location of the detected 3D keypoints are almost the same, except for a keypoint lied on the shoulder of the first frame of the 3D video sequence.

\begin{figure}[tbp]
	\centering
	\subfigure[] { \label{fig:a}    
		\includegraphics[width=1\textwidth]{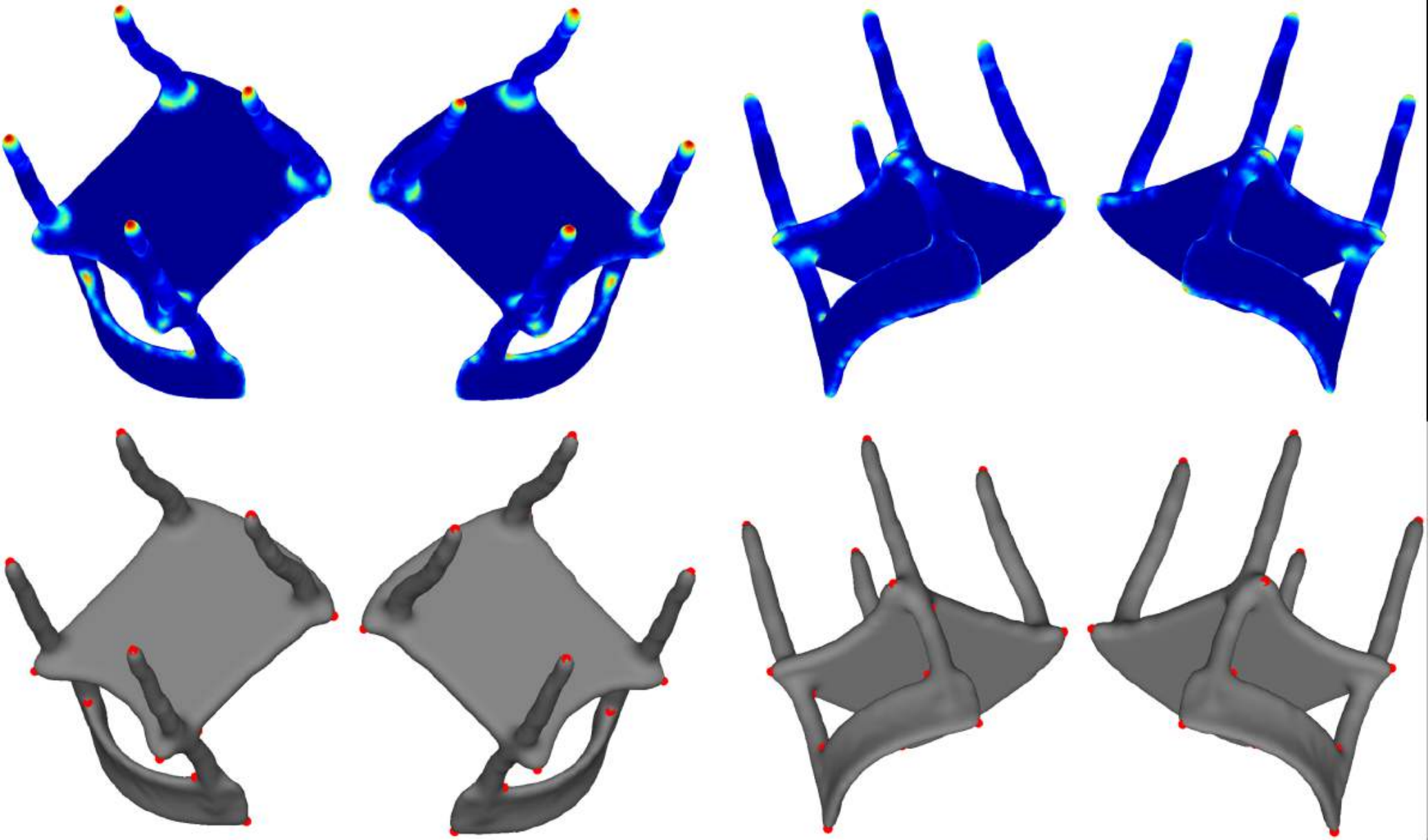}
	}   
	\subfigure[] { \label{fig:a}    
		\includegraphics[width=1\textwidth]{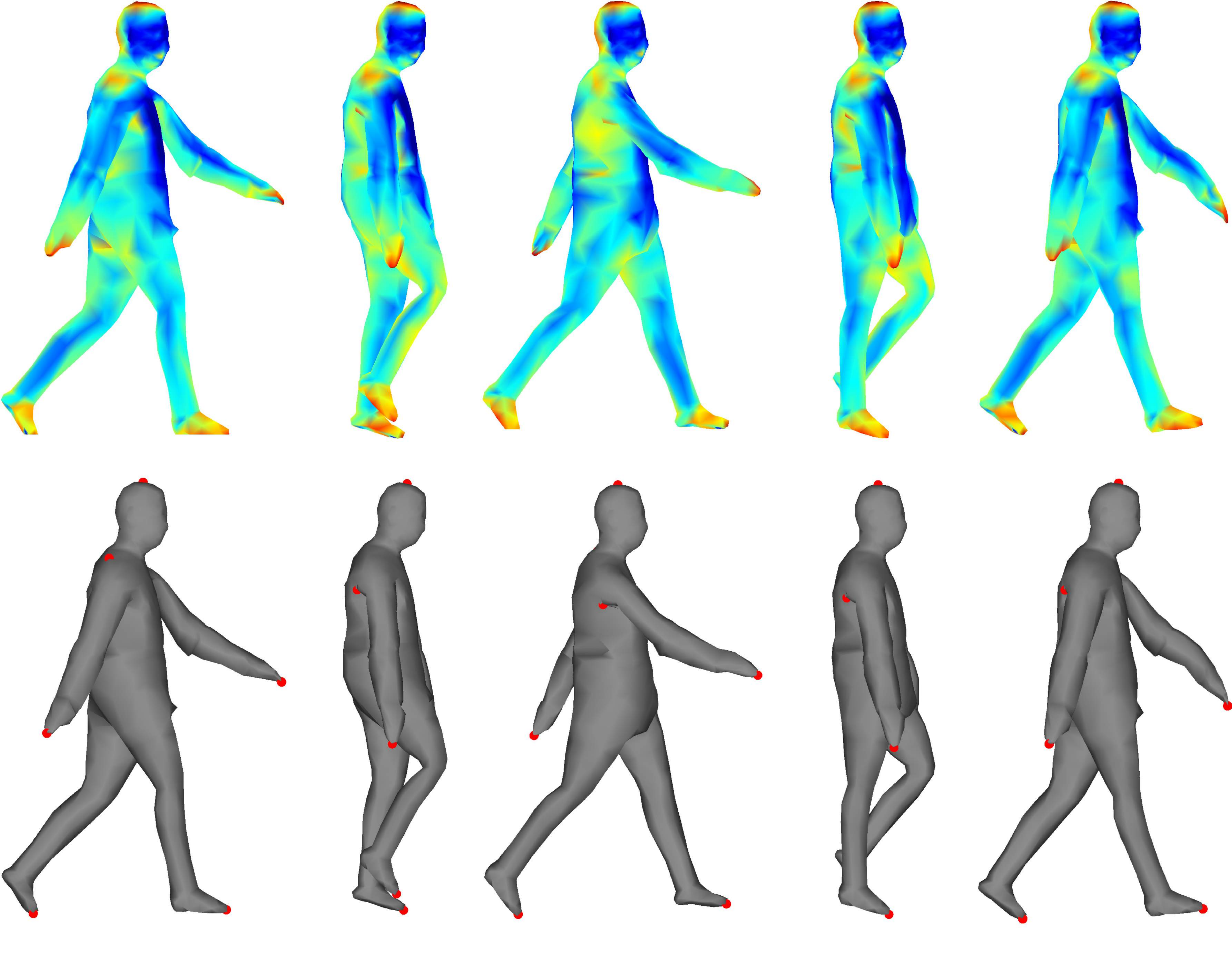}
	}
	\caption{Saliency maps (first row) and 3D keypoints (second row) detected by the proposed DNN based 3D keypoint detection algorithm.}
\end{figure}

Fig. 5 gives IOU graphs with respect to localization error tolerance \emph{$r$} for six 3D keypoint detection algorithms. We compare the performance of six methods at \emph{$n=8/\sigma=0.03$}, \emph{$n=8/\sigma=0.05$}, \emph{$n=2/\sigma=0.03$} and \emph{$n=2/\sigma=0.05$} for Dataset A in terms of IOU evaluation metrics. In the same way, we compare the performance of six methods at \emph{$n=11/\sigma=0.03$}, \emph{$n=11/\sigma=0.05$}, \emph{$n=2/\sigma=0.03$} and \emph{$n=2/\sigma=0.05$} for Dataset B in terms of IOU evaluation metrics. Our proposed approach performs best in terms of IOU evaluation metric, especially when \emph{$r$} and \emph{$n$} are relatively large.

\begin{figure}[!htb]
	\centering
	\begin{minipage}[b]{1\linewidth}
		\subfigure[3D keypoints of chair model in Dataset A detected by six algorithms] { \label{fig:a}    
			\includegraphics[width=1\textwidth]{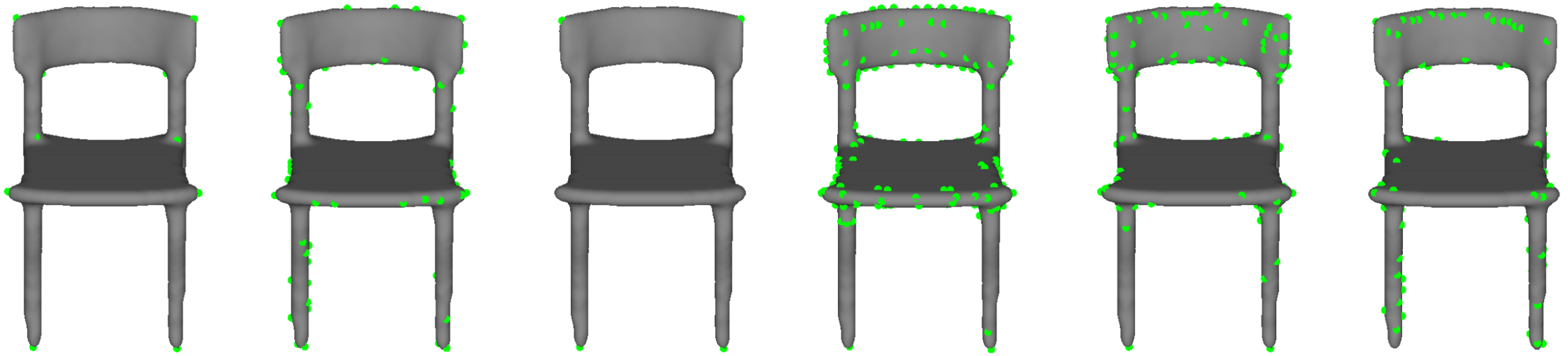}
		}   
		\subfigure[3D keypoints of armadillo model in Dataset B detected by six algorithms] { \label{fig:a}    
			\includegraphics[width=1\textwidth]{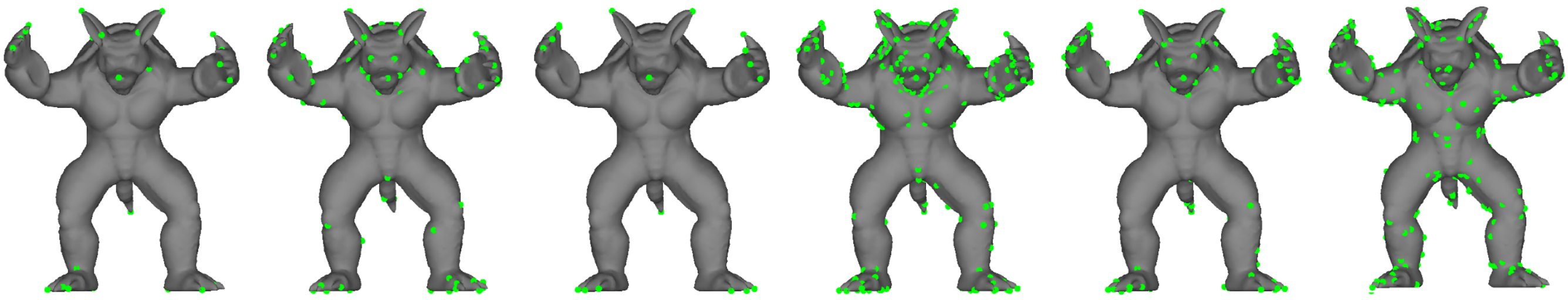}
		}
		\caption{3D keypoints detected by six algorithms. (DNN: first column; Salient points: second column; HKS: third column; Mesh saliency: fourth column; 3D-Harris: fifth column; SD corners: sixth column)}
	\end{minipage}
\end{figure}

\begin{figure}[!htb]
	\centering
	
	\includegraphics[width=0.9\textwidth]{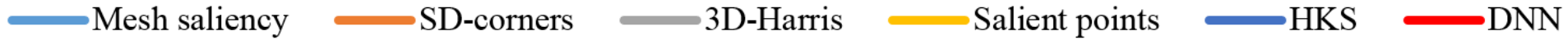}
	\begin{minipage}[b]{1\linewidth}
		\subfigure[\emph{$n=8,\sigma=0.03$} \ \ \ \ \ \emph{$n=8,\sigma=0.05$} \ \ \ \ \ \ \ \ \emph{$n=2,\sigma=0.03$} \ \ \ \ \ \ \ \ \emph{$n=2,\sigma=0.05$}] { \label{fig:a}    
			\includegraphics[width=1\textwidth]{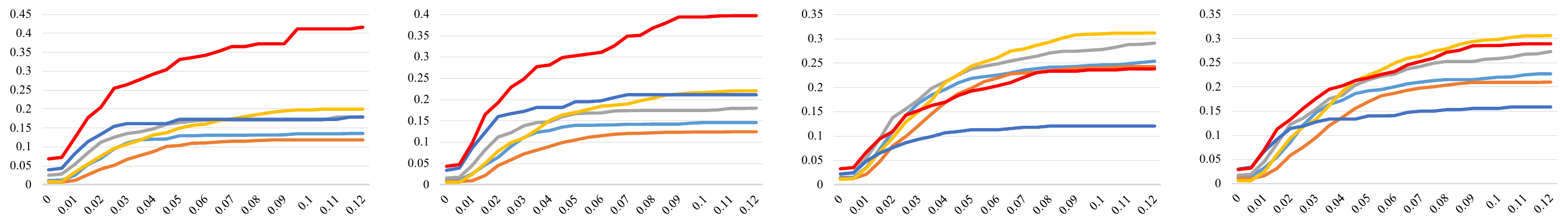}
		}   
		\subfigure[\emph{$n=11,\sigma=0.03$} \ \ \ \ \emph{$n=11,\sigma=0.05$} \ \ \ \ \ \ \emph{$n=2,\sigma=0.03$} \ \ \ \ \ \ \ \emph{$n=2,\sigma=0.05$}] { \label{fig:b}    
			\includegraphics[width=1\textwidth]{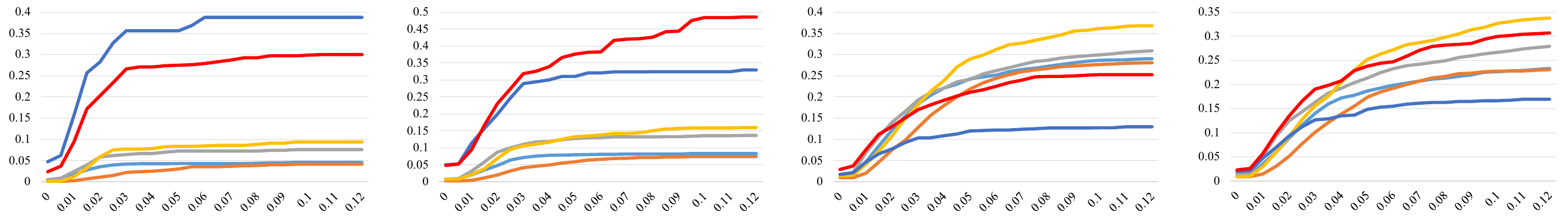}
		}
		\caption{IOU curves for Dataset A (first row) and Dataset B (second row) at variour \emph{$n/\sigma$} pairs}
	\end{minipage}
\end{figure}

To reach an overall comparison, we average the 4 types of evaluation metric scores over all settings. Fig. 6 gives average IOU, FNE, FPE and WME graphs with respect to localization error tolerance \emph{$r$} for six 3D keypoint detection algorithms, where \emph{$n \in \{2,3,...,23\} / \sigma \in \{0.01,0.02,...,0.1\}$} is for Dataset A and \emph{$n \in \{2,3,...,16\} / \sigma \in \{0.01,0.02,...,0.1\}$} is for Dataset B. Digital results over all settings are summarized in Table 4. From Fig .6 and Table 4, we can see that our proposed approach performs best, especially when localization error tolerance \emph{$r$} is relatively large.

\begin{figure}[!htb]
	\centering
	
	\includegraphics[width=0.9\textwidth]{label}
	\begin{minipage}[b]{1\linewidth}
		\subfigure[4 kinds of performance curves on Dataset A, \emph{$n\in \{2,3,...,23\},\sigma \in \{0.01,0.02,...,0.1\}$}] { \label{fig:a}    
			\includegraphics[width=1\textwidth]{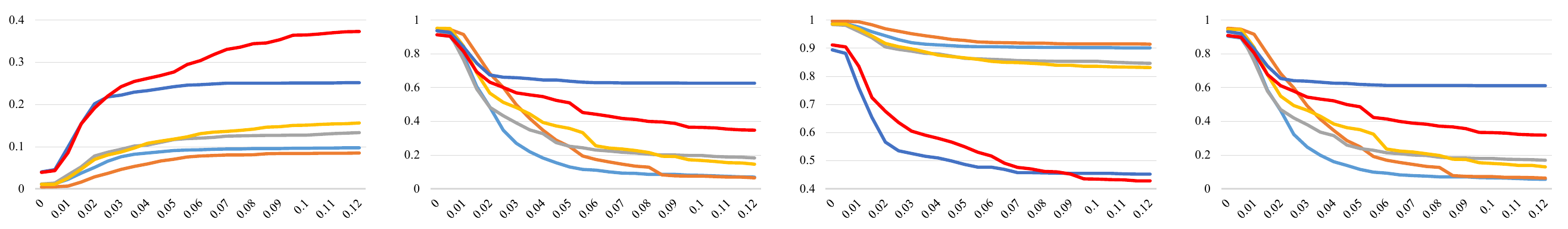}
		}   
		\subfigure[4 kinds of performance curves on Dataset B, \emph{$n \in \{2,3,...,16\},\sigma  \in \{0.01,0.02,...,0.1\}$}] { \label{fig:b}    
			\includegraphics[width=1\textwidth]{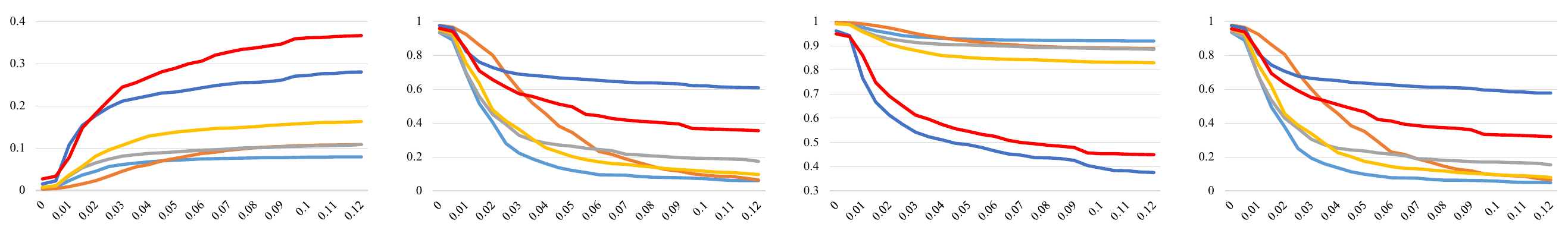}
		}
		\caption{Four kinds of performance curves on Dataset A and B. IOU (first column), FNE (second column), FPE (third column), WME (fourth column)}
	\end{minipage}
\end{figure}

\begin{table}[!htb]  
	\centering
	\renewcommand\arraystretch{1.2}
	\caption{Average IOU, FNE, FPE, WME on Test Dataset A (\emph{$n \in \{2,3,...,23\}, \sigma \in \{0.01,0.02,...,0.1\}$}) and B (\emph{$n \in \{2,3,...,16\}, \sigma \in \{0.01,0.02,...,0.1\}$})}
	\begin{tabular}{ccccccccc}  
		\hline
		\cline{1-9}
		&IOU-A  &FNE-A  &FPE-A  &WME-A               \ &IOU-B &FNE-B &FPE-B &WME-B \\ \hline  
		Mesh saliency &0.078 &0.248 &0.919 &0.232                &0.063 &0.225 &0935 &0.210\\        
		SD-corners &0.061 &0.328 &0.936 &0.326                   &0.072 &0.379 &0.924 &0.380\\        
		3D-Harris &0.102 &0.343 &0.879 &0.329                    &0.084 &0.330 &0.910 &0.307\\        
		Salient points &0.111 &0.376 &0.875 &0.361               &0.122 &0.295 &0.868 &0.274\\
		HKS &0.216 &0.671 &0.530 &0.656                          &0.218 &0.686 &0.519 &0.662\\
		\textbf{DNN} &\textbf{0.275} &\textbf{0.509} &\textbf{0.561} &\textbf{0.484}                          &\textbf{0.272} &\textbf{0.516} &\textbf{0.578} &\textbf{0.490}\\
		\hline
		\cline{1-9}
	\end{tabular}
\end{table}

\section{Conclusion}
In this paper, we propose a new 3D keypoint detection algorithm on the basis of deep learning by formulating the 3D keypoint detection as a regression problem using DNN with SAE as our regression model. It's the first time that DNN with SAE has been used to detect 3D keypoints. Both local information and global information of a 3D mesh model in multi-scale space are fully utilized to detect whether a vertex is a keypoint or not. Three tpyes of geometric properities of surface of a 3D mesh model are used to formulate the local information: 1) the Euclidean distance between neighborhood rings to the tangent plane; 2) the angle of normal vector between the vertex and its neighborhood rings; 3) various curvatures. For global information, we consider the properties of log-Laplacian spectrum of a 3D mesh model used in \cite{song2014mesh}. SAE can effectively extract the internal structure of these two kinds of information and formulate high-level features for them, which is beneficial to the regression model. Three SAEs are used to formulate the hidden layers of the DNN and then a logistic regression layer is trained to process the high-level features produced in the third SAE. These four layers are stacked together to formulate a DNN as the regression model of our 3D keypoint detection algorithm. 

Experimental results indicate that the proposed DNN based 3D keypoint detection algorithm outperforms other five state-of-the-art methods in terms of IOU metric, especially when localization tolerance error \emph{$r$} is relatively large. Besides, the 3D keypoints detected by our approach are stable and more in accord with human visual characteristics.

\bibliographystyle{splncs}
\bibliography{egbib}
\end{document}